\documentclass[letterpaper, 10 pt, conference]{ieeeconf}  

\IEEEoverridecommandlockouts                              

\usepackage{graphicx}
\usepackage{amsmath}
\usepackage[hidelinks]{hyperref}
\usepackage{units}
\newcommand{\V}[1]{\mathbf{#1}} 

\usepackage[utf8]{inputenc}
\usepackage{todonotes}
\usepackage{breqn}
\usepackage{amsfonts}
\usepackage{times}
\usepackage{graphicx}
\usepackage{todonotes}
\usepackage[hidelinks]{hyperref}
\usepackage[ruled,vlined]{algorithm2e}  
\usepackage{algpseudocode}
\usepackage{cite}

\algnewcommand\algorithmicforeach{\textbf{for each}}
\algnewcommand\algorithmicor{\textnormal{\textbf{r} }}
\algdef{S}[FOR]{ForEach}[1]{\algorithmicforeach \ #1 \ \algorithmicdo}
\SetKwFor{Loop}{Loop}{}{EndLoop}

\usepackage{psfrag}
\usepackage{pstool}

\usepackage{dblfloatfix} 
\usepackage{float} 
\usepackage{capt-of}

\title{\Large \bf
Dynamics of spherical telescopic linear driven rotation robots
}

\author{Jasper Zevering*$^{1}$,  Dorit Borrmann,$^{2}$ Anton Bredenbeck$^{3}$, Andreas Nuechter$^{1}$
        \thanks{ *\quad  \tt\small jasper.zevering@uni-wuerzburg.de}%
\thanks{$^{1}$ Informatics VII : Robotics and Telematics - University of Wuerzburg}%
\thanks{$^{2}$ Technical University of Applied Sciences Schweinfurt}%
\thanks{$^{3}$ TU Delft}%
\thanks{We acknowledge funding from the ESA Contract No. 4000130925 /20/NL/GLC  for the ``DAEDALUS -- Descent And Exploration in Deep Autonomy of Lava Underground Structures'' Open Space Innovation Platform (OSIP) lunar caves-system study and the Elite Network Bavaria (ENB) for providing funds for the academic program ``Satellite Technology''. \vspace{-0.5cm}}
}

\begin{document}

\maketitle
\thispagestyle{empty}
\pagestyle{empty}

\begin{abstract}
Lunar caves are promising features for long-term and permanent human presence on the moon. 
However, given their inaccessibility to imaging from survey satellites, the concrete environment within the underground cavities is not well known. 
Thus, to further the efforts of human presence on the moon, these caves are to be explored by robotic systems. 
However, a set of environmental factors make this exploration particularly challenging. 
Among those are the very fine lunar dust that damages exposed sensors and actuators and the unknown composition of the surface and obstacles within the cavity. 
One robotic system that is particularly fit to meet these challenges is that of a spherical robot, as the exterior shell completely separates the sensors and actuators from the hazardous environment.
This work introduces the mathematical description in the form of a dynamic model of a novel locomotion approach for this form factor that adds additional functionality.
A set of telescopic linearly extending rods moves the robot using a combination of pushing away from the ground and leveraging the gravitational torque. 
The approach allows the system to locomote, overcome objects by hoisting its center of gravity on top, and transform into a terrestrial laser scanner by using the rods as a tripod. 
\end{abstract}


\begin{figure}[b]
  \begin{minipage}{\textwidth}
    \centering
    \includegraphics[width=\textwidth]{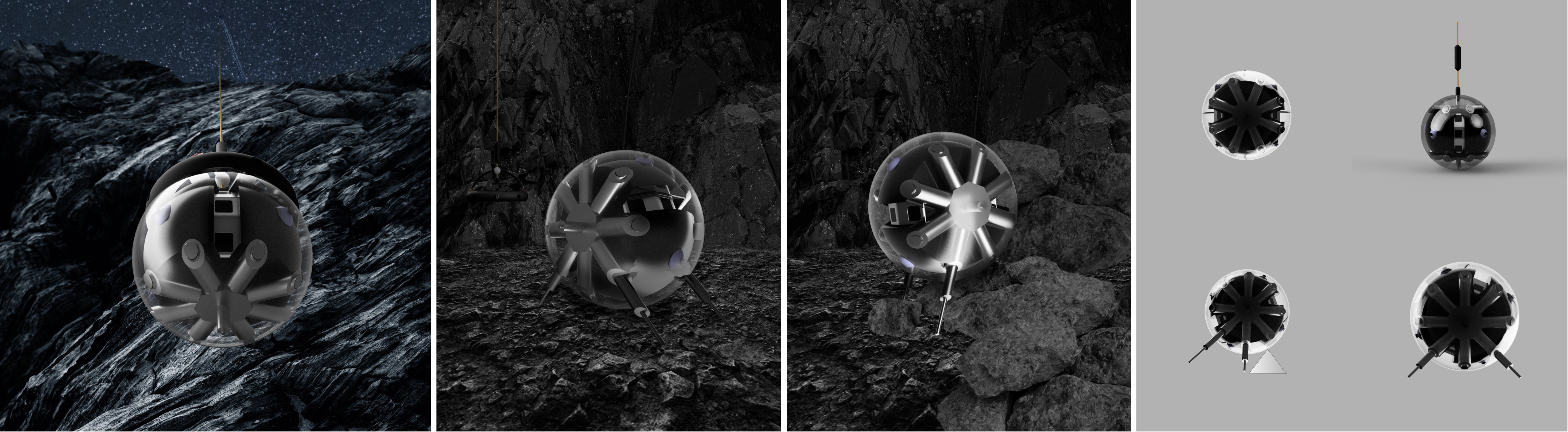}
    \vspace{-0.5cm}
    \caption{The DAEDALUS sphere. From left to right: First, Daedalus sphere is descended into the pit by a crane. Second, DAEDALUS in scanning mode. Third, Different modes of DAEDALUS. Fourth, DAEDALUS overcoming obstacles by pushing with its rods.}
    \label{fig:DaedalusCollage}
  \end{minipage}
\end{figure}%


\section{Introduction}

Establishing a permanent human presence on celestial bodies has long been a goal of humanity's space exploration efforts.
It will enable a vast amount of research into humanity's origin and function as a hub for further exploration into deeper space, such as missions to Mars.
 More recently, NASA formulated this more concretely within the Artemis program.
One of the mission goals is to enable extended human presence on the moon~\cite{russell2014artemis}.
Good locations for long to permanent human settlements on the moon are lunar caves. They provide shelter from radiation and maintain a more consistent and comfortable temperature than the lunar surface.
However, barring the areas within the field of view of survey satellites, little is known about the regions, especially the underground tunnels, which are of great interest.
In order to survive in this harsh and mostly unknown environment and complete an exploration task, a robotic system must abide by several requirements. 
Experts expect that the individual rocks are very sharp and that the surface dust -- so-called regolith -- is very fine due to the lack of deterioration effects, such as wind~ \cite{mckay1991lunar,gies1996effect}. 
Both pose a danger to sensors and actuators, e.g., by scratching the lenses of optical sensors or causing friction in gears and motors. 
Hence, any robotics system must mitigate these damaging environmental effects by sealing the sensors and actuators from these hazards.   
Further, the significant uncertainty associated with the environment within the tunnels requires a versatile locomotion system. 
In particular, it must be capable of overcoming obstacles of various sizes.
Overcoming objects of unknown size is non-trivial for conventional wheeled robotic systems, where the wheels fixed size is directly proportional to the size of permissible objects.

\enlargethispage{-6.5cm}

To kickstart the research into robotic exploration systems suited for these requirements, the European Space Agency (ESA) has called for design proposals for robotic platforms capable of handling the various challenges posed by the tunnels.  
One design, as shown in figure~\ref{fig:DaedalusCollage}, deemed particularly fit to tackle these challenges is that of a spherical robot as introduced in the DAEDALUS project~\cite{rossi2021daedalus}.
Given the complete separation between the interior from the hazardous environment as well as its omnidirectional nature, a spherical robot is a promising design choice~\cite{crossley2006literature}.
After being hoisted into the cave by a crane system, the exterior shell protects the sensitive sensors and the various actuators from the fine regolith dust and impacts from traversing over the rough terrain.
Besides multiple sensors used for creating a three-dimensional model of the environment, the spherical robot houses a novel locomotion approach:  by extending and retracting a set of rods -- each on a linear axis --  the robot moves through its environment via a combination of pushing away from the ground and leveraging the gravitational torque imposed by an extended rod.
While this approach is more complex than the traditional spherical robots drive systems, it promises to solve multiple issues. 
In addition to basic locomotion, this locomotion approach allows the robot to traverse obstacles by hoisting its center of gravity (CoG) onto the obstacles and passing through uneven terrain by adjusting the rod length to a local environment map. 
Further, the rods provide a stable tripod, effectively transforming the mobile robot into a terrestrial laser scanner, thus facilitating exact three-dimensional models of the environment. 

This work introduces the mathematical description of a spherical robot equipped with the above locomotion system in the form of a dynamic model. 
Further, it places this novel locomotion approach into context with other approaches for the locomotion of spherical robots.

\section{Related Work}

The vast majority of spherical robots use pendulum drives \cite{chase2012review, koshiyama1993design, BYQ-III, alizadeh2011quadratic, pokhrel2013design, moazami2019kinematics, yang2020design, mahboubi2013design, yoon2011spherical, kaznov2010rotundus}, weight shifting approaches \cite{mojabi2002introducing, mukherjee1999simple,lux2005alternative,zhao2012mechanical} or internal drive units (IDUs) \cite{ylikorpi2007ball, halme1996motion, alves2003design, zhan2011design, chen2013design, nguyen2018virgo, karavaev2015dynamics, karavaev2016nonholonomic, reina2004rough, qiang2008back}.
Internal generation of momentum is also used in many cases \cite{bhattacharya2000spherical, luna, chemel1999cyclops, qingxuan2009motion}
and sometimes deformation of the robot \cite{armour2007jumping, sugiyama2004crawling, kim2016hopping, chen2017soft, agogino2018super, wait2010self}.

Previous work in the field of spherical robots using linear actuators extrinsic to the shell is limited.
In \cite{Ocampa2014} Ocampo-Jimenez and his team investigated and described the approach of a internal pendulum and additionally using small linear actuators for lifting the sphere a little in case of it getting stuck.
The performed simulation showed the capability of freeing the sphere when it is stuck between obstacles up to $\frac{1}{8}$ of the height of the robot in comparison to $\frac{1}{10}$ without the usage of the actuator.
This leaves room for improvement, also the actuators were not bind into the locomotion itself.
Kim et al. in 2010~\cite{kim2010kisbot} has also introduced a prototype using a linear actuator, named Kisbot.
It consists of a sphere divided into three parts: one middle ring and two outer semispheres.
The rotatable semispheres are mounted onto the middle part and are actuated, and have a linear extendable part.
For locomotion, it has two driving modes.
One is the pendulum-driven rotation,  using the non-homogeneous mass distribution of the rotatable  semispheres.
The second mode is the wheeling mode, where both extendable parts are extended, resulting in the overall functioning of the robot like a one-wheel car.
Nonetheless, the linear extension is also used for locomotion as it allows the sphere to push itself on top of obstacles.
Further, extending these parts leads to an abrupt stop, if extended on the side where the robot is rolling towards.
Both driving modes were tested and evaluated.
Regarding lifting and stopping, the concept was only described and not implemented.
The German company Festo AG \& Co. KG created BionicWheelBot, that imitates the so-called flic-flac spider~\cite{BionicWheelBot} and uses its leg setup for walking and rolling.
While rolling the round structure itself consists of six legs of the robot, the other two are used for pushing.
While walking only the six legs are used for the walking itself as the two, which push during rolling, cannot be used due to mechanical restrictions.
The pushing legs have a joint that enables pushing despite a lateral orientation.
This way of pushing for rotation comes very close to pushing with linear actuators.
As only one pair of legs is used for pushing, this leads to a varying rotational speed during one rotation.


\section{Physical Representation}
\label{subsec:physicalRepresentation}
We now want to evaluate the fundamental physical interactions of the TLDR robot.
This evaluation is based on \cite{hermans1995symmetric,  khaleghian2017technical, pinto2001rolling, bowyer2014dynamic}, and \cite{hierrezuelo1995sliding}, which mostly deal with the integration of friction into the analysis of a rolling spheroid.
This physical evaluation will help identify the limitations of the prototype and its requirements.
In the following, we will evaluate the pushing approach with and without slip as well as the leverage approach in terms of their physical representation.
\subsection{Pushing with no slip}
\begin{figure}
    \centering
    \includegraphics[width = 0.7 \linewidth]{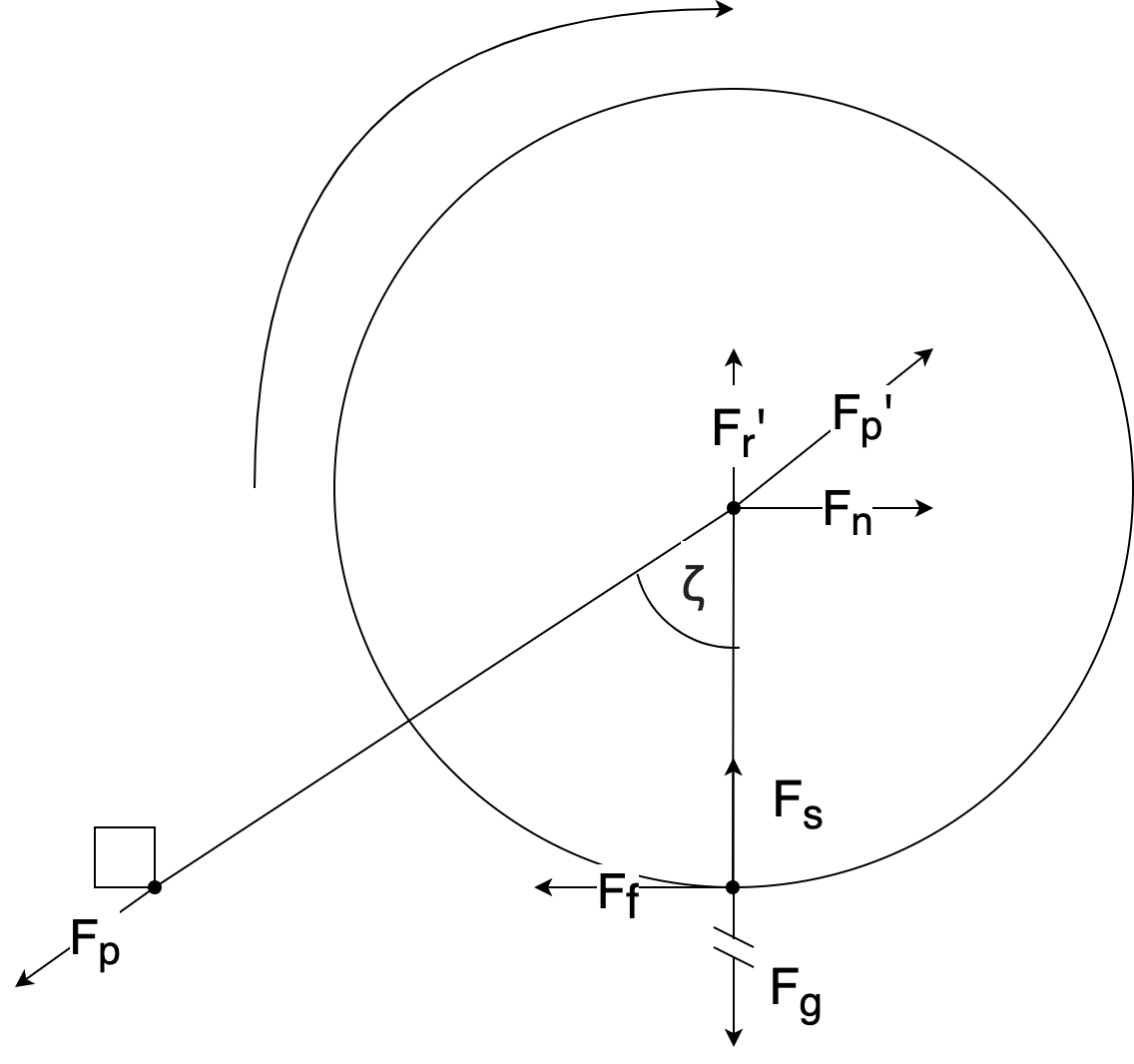}
    \caption{Force evaluation of the pushing approach with no slip due to an obstacle.}
    \label{fig:PhyiscalPushingFixedPole}
\end{figure}

First, we look at the simplest case: pushing with no slip at the pole end.
Here, we assume an obstacle at the end of the actively pushing pole.
This prohibits every slip of the pole.
Figure~\ref{fig:PhyiscalPushingFixedPole} shows the simplified forces in this case.
We focus on the direct force interaction of the pole and the robot.
Forces like air resistance, sinking in due to the softness of the ground, etc., are neglected.
As the pole pushes with $\V F_P$ against the ground and non-moving obstacle, the counterforce $\V F_p^{'}$ acts at the middle of the robot and has the two components $\V F_r$ and $\V F_n$, which can be calculated by 
\begin{equation}
   \V F_p^{'}  =  F_p^{'} \cdot
    \begin{bmatrix}
    \sin(\zeta) \\ \cos(\zeta)
    \end{bmatrix} =  
    \begin{bmatrix}
     F_r^{'} \\  F_n
    \end{bmatrix}
    \,.
\end{equation}
The vertical $\V F_r$ is directly countered by the gravity $\V F_g$.
The difference between $\V F_g$ and $\V F_r$ is compensated by the structural force $\V F_s$.
If $\V F_r$ is larger than $\V F_g$, the sphere starts to accelerate in the vertical direction.
The horizontal force $\V F_n$ is countered by the frictional force $\V F_f$.
Let $\mu_{rs}$ be the sum of the friction coefficient $\mu_s$ of the robot and surface, and the rolling friction coefficient $c_r$, which depends on how much the specific structure sinks into the surface due to its own weight. Then, $\V F_f$ is defined as
\begin{equation}
    \V F_f = \mu_{rs} \cdot (- \V F_n) \,.
\end{equation}
Note that $\mu_{rs}$ is chiefly determined experimentally.
We will often assume it to be 1, which means a good grip with no slip.
The difference between $\V F_f$ and $\V F_n$, which is always in the direction of $\V F_n$ as $\mu_{rs}$ is between zero and one, leads to the translation of the sphere.
Let $a_{\text{direct}}$ be the acceleration of this direct translation without translation, and m be the mass of the robot, then
\begin{equation}
    a_{\text{direct}} = \frac{F_n - F_f}{m}=  \frac{F_n \cdot(1-\mu_{rs}) }{m}
    \label{Eq:aDirectObstacle}
\end{equation}
Let $\V \tau_f$ be the torque generated by $\V F_f$, and $r$ be the position vector of the acting point of $\V F_f$, then
\begin{align}
   \V \tau_f = \V r \times \V F_f \\
   \tau_f = r_m \cdot F_f \,.
   \label{Eq:tauF}
\end{align}
Let $I$ be the moment of inertia of the sphere and $\dot{\omega}$ the acceleration of the angular velocity, then
\begin{equation}
\dot{\omega} = \frac{\tau_f}{I}\,.
\end{equation}
$I$ depends on the mass, shape, and mass distribution of the robot.
For a massive sphere, this is $\frac{2}{5}m r^2$, but the length of extension of the pole needs to be considered as it changes the moment of inertia, even in this case.
Also, if the robot is not built perfectly balanced, $I$ has different values for each rotation axis.
Therefore, it actually is a tensor, i.e., 3x3 matrix.
With mechanical simulations or experiments, this tensor is determinable.
We will refer to it as general $I$.

\begin{figure}
    \centering
\includegraphics[width =  \linewidth]{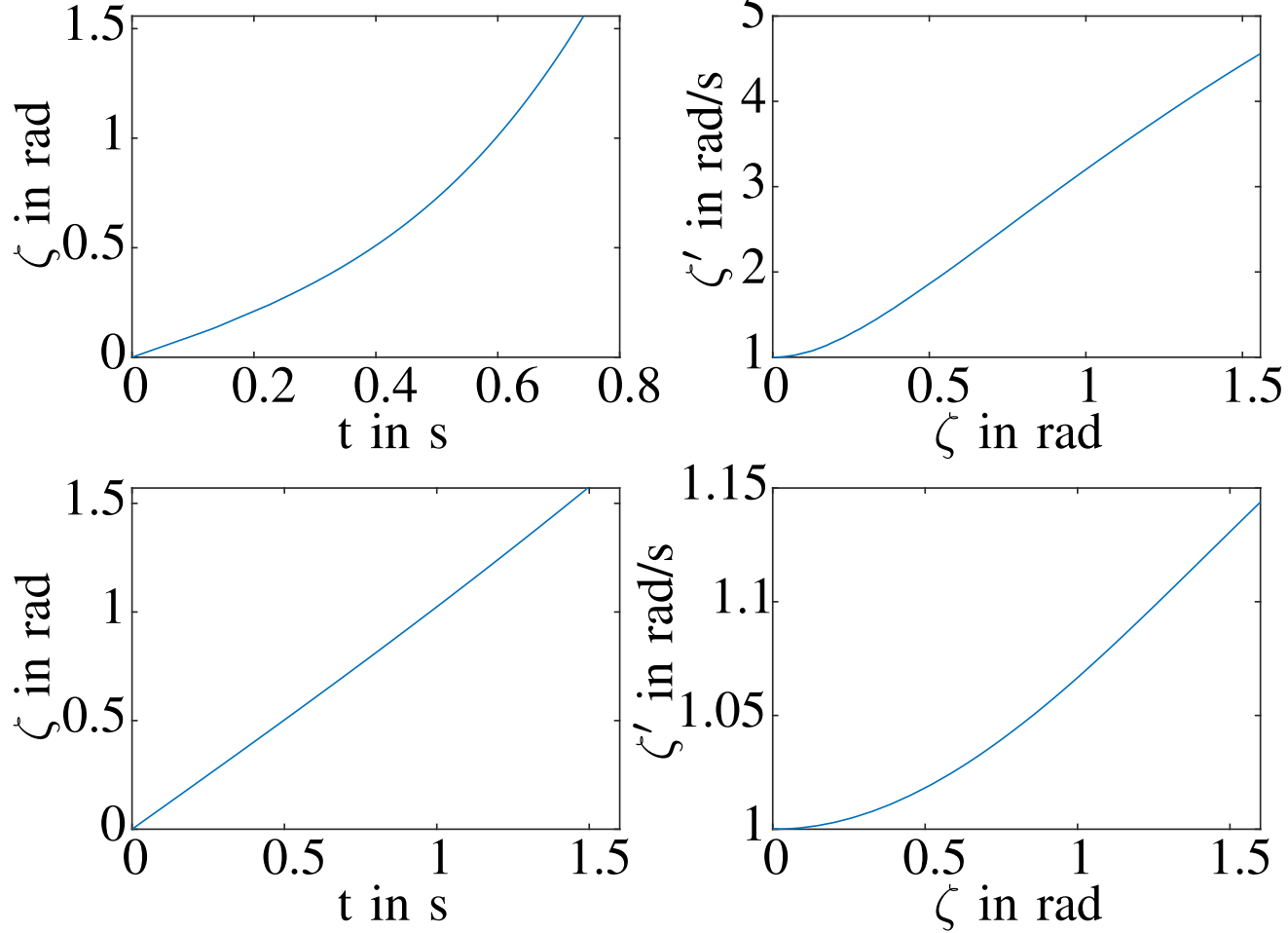}
    \vspace{-0.5cm}
    \caption{Analytical solution of Equation~\eqref{Eq:aDotOmegaNoSlip}. Above: Solution for $A=10$. Below: Solution for $A=0.15$. }
    \label{fig:DifferentialSolution1}
\end{figure}

The initiated rotation and the translation by sliding cause the rolling of the robot.
The translation consists of not only the direct linear slip but also the rotation-initiated translation in the form of 
\begin{equation}
    a_{\text{rotation}} = 2 \pi \cdot r_m \cdot \dot{\omega}\,.
    \label{Eq:aRotationObstacle}
\end{equation}
This leads to the overall system.
Let $a$ be the overall acceleration of the robot in the vertical direction (so $a_{direct}+a_{rotation}$), $\dot{\omega}$ the angular acceleration, and $m$ the mass of the robot, then

\begin{equation}
    \begin{bmatrix}
    a \\ \dot{\omega}
    \end{bmatrix}
    = 
    \begin{bmatrix}
    \frac{\sin(\zeta)\cdot F_p \cdot(1-\mu_{rs}) }{m} + 2 \pi \cdot r_m^2 \cdot \mu_{rs} \cdot \sin(\zeta)\frac{F_p}{I}\\ r_m \cdot \mu_{rs} \cdot \sin(\zeta) \cdot \frac{F_p}{I}
    \end{bmatrix}\,.
\end{equation}
For the case where there is no slip, $\mu_{rs} = 1$, so we get
\begin{align}
 \begin{bmatrix}
    a \\ \dot{\omega}
    \end{bmatrix}
    &= 
    \begin{bmatrix}
    2 \pi \cdot r_m^2 \cdot \sin(\zeta) \cdot \frac{F_p}{I}\\ r_m  \cdot \sin(\zeta) \cdot \frac{F_p}{I}
    \end{bmatrix} \nonumber
   \\  &= 
    \begin{bmatrix}
    2 \pi \cdot r_m \cdot \dot{\omega}\\ r_m  \cdot \sin(\zeta) \cdot \frac{F_p}{I}
    \end{bmatrix}\,.
    \label{Eq:aDotOmegaNoSlip}
\end{align}
Note that as $\dot{\omega} = \ddot{\zeta}$, this is a second-order nonlinear differential equation.
For the evaluation, we assume $\frac{r_m\cdot F_p}{I}$ to be constant and refer to it as constant $A$.
Solving it analytically gives the result shown in Figure~\ref{fig:DifferentialSolution1}, once for a high factor $A=10$ and once for the factor $A=0.15$, which is the magnitude of the later-introduced prototype.

\begin{figure}
    \centering
    \def\svgwidth{\linewidth}
\begingroup%
  \makeatletter%
  \providecommand\color[2][]{%
    \errmessage{(Inkscape) Color is used for the text in Inkscape, but the package 'color.sty' is not loaded}%
    \renewcommand\color[2][]{}%
  }%
  \providecommand\transparent[1]{%
    \errmessage{(Inkscape) Transparency is used (non-zero) for the text in Inkscape, but the package 'transparent.sty' is not loaded}%
    \renewcommand\transparent[1]{}%
  }%
  \providecommand\rotatebox[2]{#2}%
  \newcommand*\fsize{\dimexpr\f@size pt\relax}%
  \newcommand*\lineheight[1]{\fontsize{\fsize}{#1\fsize}\selectfont}%
  \ifx\svgwidth\undefined%
    \setlength{\unitlength}{193.07740281bp}%
    \ifx\svgscale\undefined%
      \relax%
    \else%
      \setlength{\unitlength}{\unitlength * \real{\svgscale}}%
    \fi%
  \else%
    \setlength{\unitlength}{\svgwidth}%
  \fi%
  \global\let\svgwidth\undefined%
  \global\let\svgscale\undefined%
  \makeatother%
  \begin{picture}(1,0.92521895)%
    \lineheight{1}%
    \setlength\tabcolsep{0pt}%
    \put(0,0){\includegraphics[width=\unitlength]{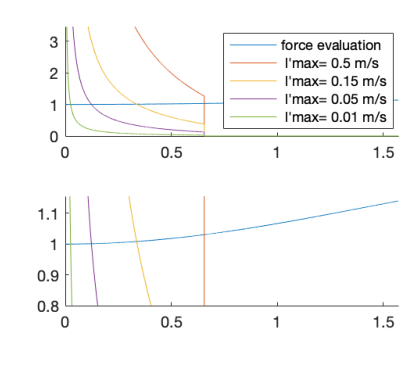}}%
    \put(0.48,0.048){\makebox(0,0)[lt]{\lineheight{1.25}\smash{\begin{tabular}[t]{l}$\zeta$ in rad\end{tabular}}}}%
    \put(0.07534384,0.22){\rotatebox{90}{\makebox(0,0)[lt]{\lineheight{1.25}\smash{\begin{tabular}[t]{l}$\zeta '$ in rad\end{tabular}}}}}%
    \put(0.0946738,0.65){\rotatebox{90}{\makebox(0,0)[lt]{\lineheight{1.25}\smash{\begin{tabular}[t]{l}$\zeta '$ in rad\end{tabular}}}}}%
    \put(0.48,0.49){\makebox(0,0)[lt]{\lineheight{1.25}\smash{\begin{tabular}[t]{l}$\zeta$ in rad\end{tabular}}}}%
  \end{picture}%
\endgroup%

\vspace{-0.8cm}
    \caption[Analytical solution of Equation~\eqref{Eq:aDotOmegaNoSlip} in comparison to the geometrically possible solution.]{Analytical solution of Equation~\eqref{Eq:aDotOmegaNoSlip} with $A=0.15$ and the geometrical possible $\dot{\zeta}$s with a radius of \unit[0.4]{m} and a maximum length of \unit[0.1]{m}. The force solution is limited by either the extension speed or the sheer fact that the pole length is not long enough to reach $\zeta$}
    \label{fig:ForceVsGeometricalLimitaion}
\end{figure}
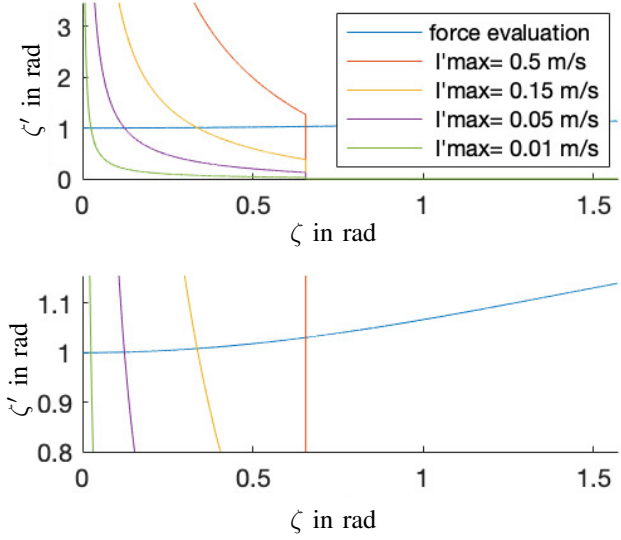

As expected, with an increasing $\zeta$, the angular velocity also raises.
The plot cannot be seen as a direct representation of the actual $\zeta$ as this implies an obstacle that is at the assumed distance, which moves with the extending pole and counters the force perfectly.
This is why we assumed a linear $\omega$ for the mathematical representation.
However, even if that case occurs, the resulting angular velocity has a limit, defined by the pole length and extension speed.
Comparing this limit with the solution of the force equation show that the theoretical introducible speed as per the force calculation exceeds what is geometrically possible.
Figure~\ref{fig:ForceVsGeometricalLimitaion} visualizes this.

The larger $\zeta$ becomes, the slower the maximum $\dot{\zeta}$ can be up to the point where the maximum possible length of the pole $l_{\text{max}}$ limits the reachable $\theta$.
In the shown calculation, the configuration with the fastest $\dot{l}$ does indeed cut the force calculation at the vertical transition to \unit[0]{rad}, which means that in this case, the length was the limit.
Combining the geometrical and physical behaviors shows the overall problem of the system.
These both influences do not just limit each other or reinforce each other.
The symbiosis of both leads to a system that is hard to predict.
The following are two exemplary reasons for it:
\begin{itemize}
    \item A pole jumps over the ground until it finds an obstacle or reaches a certain level of bending. 
\item A rotation that leads to an extension of the pole in the air without ground contact initiates leverage in the other direction than intended.
\end{itemize}
Because of these, the starting $\omega$ of each pole at $\zeta= \alpha$ will always be different.

Overall, this thesis will not provide a complete solution to the whole system.
It focuses on evaluations for general limitations and dimensioning components.
Therefore, the further evaluations and their resulting differential equations will not be solved as the acceleration at a given $\zeta$ is the only value of interest.

\subsection{Pushing with slip}
\begin{figure}[ht]
    \centering
    \includegraphics[width = 0.7\linewidth]{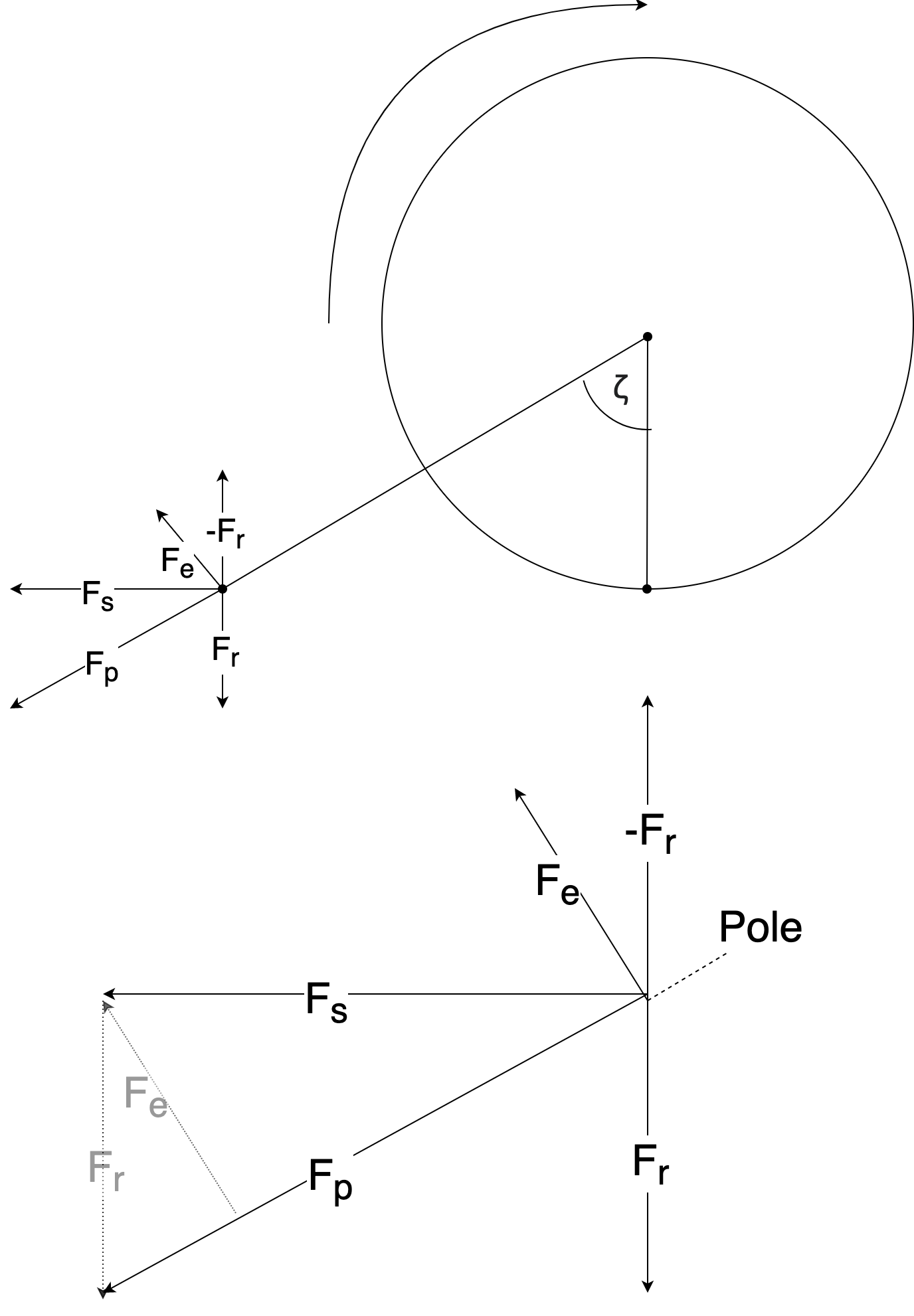}
    \caption{Force evaluation of the pushing approach with a complete slip of the poles.}
    \label{fig:PhyiscalPushingNoFixedPole}
\end{figure}

\begin{figure}
    \centering
    \includegraphics[width = 0.8 \linewidth]{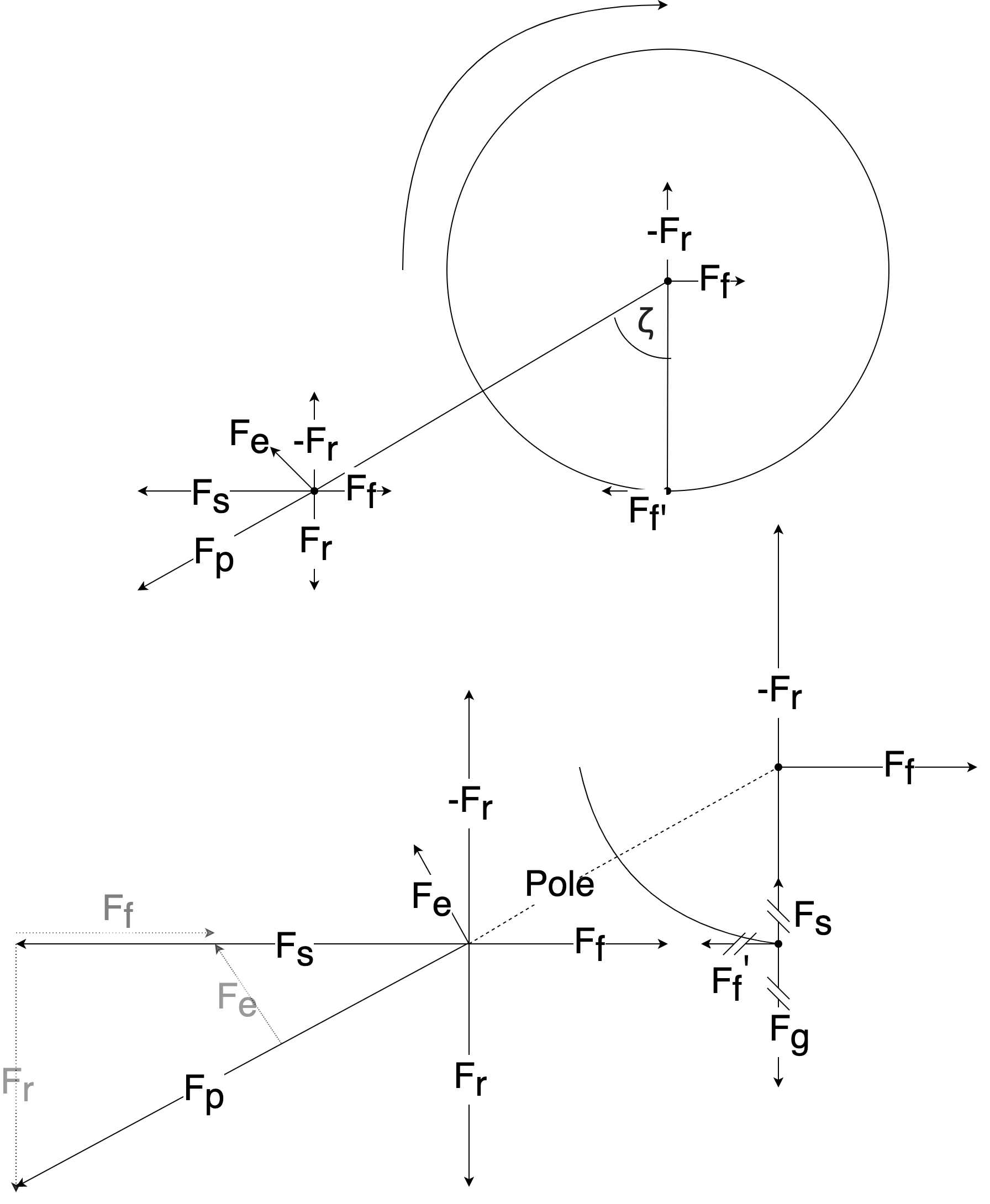}
    \caption{Force evaluation of the pushing approach with variable friction applied at the pole ends.}
    \label{fig:PhyiscalPushingFinal}
\end{figure}

The subsequent evaluation is for the situation in which there is no obstacle at the end of the pole.
This not only changes the counterforce from the pole applied on the sphere but also another component, the lever.
This does not mean it is the leverage approach, where the poles are extended on the side of the robot towards which it is to roll.
This is evaluated separately in subsection~\ref{subsec:Physicalleverage}.
However, we consider a ground with no friction so that the pole will extend and touch the ground; however, rather than that the applied force being directly countered by the ground, the pole slips to the side.
This is comparable to a person trying to push himself/herself while standing on ice only using a pole with a big flat surface.
When trying to push himself/herself, the pole will slide over the ice surface, but the person will not move.
The sphere generates no translation on a friction-less surface, but it has the ability to initiate rotation.
Figure~\ref{fig:PhyiscalPushingNoFixedPole} illustrates the working forces for this.
Like before, the starting point is the force of the pole $\V F_p$.
It pushes into the ground at the angle $\zeta$, which splits it up into parts, the force in the slip direction $\V F_s$ and the force toward the ground $\V F_r$.
$\V F_p$ splits as
\begin{equation}
   \V F_p  =  F_p \cdot
    \begin{bmatrix}
    \cos(\zeta) \\ \sin(\zeta)
    \end{bmatrix} =  
    \begin{bmatrix}
     F_r \\ F_s
    \end{bmatrix}
    \,.
    \label{Eq:F_pSplitUp}
\end{equation}
Like with the obstacle, $\V F_r$ is countered by the ground itself, producing $ -\V F_r$.
The difference is the arising $\V F_s$, which was previously countered by the obstacle, hence becoming $\V F_n$, but it now has no counterpart.
This creates an acceleration of the pole tip in this direction.
However, the point is part of a whole structure, and because of this, $\V F_s$ results in the lever force, perpendicular to the line of the pole, $\V F_e$.
The amount is determined by 
\begin{equation}
    F_e = \cos(\zeta) \cdot F_s
\end{equation}
Let $\tau_e$ be the torque generated by force $\V F_e$,  then
\begin{align}
   \V \tau_e = \V r \times \V F_e \\
   \tau_e = (l(\zeta) + r_m) \cdot F_e \,.
   \label{Eq:tauE}
\end{align}
Let $I$ be the moment of inertia of the sphere and $\dot{\omega}$ the acceleration of the angular velocity, then
\begin{equation}
\dot{\omega} = \frac{\tau_e}{I} =   (l(\zeta) + r_m) \cdot F_e \cdot I^{-1} \,.
\end{equation}
For $l(\zeta)$, we can use Equation~\eqref{Eq:lWithZeta}.
This leads to 
\begin{equation}
\begin{bmatrix}
a\\ \dot{\omega} 
\end{bmatrix}
= 
\begin{bmatrix}
0
\\
 (\frac{r_m}{\cos(\zeta)}) \cdot F_e \cdot I ^{-1}
\end{bmatrix}
\,.
\label{Eq:aDotOmegaWithSlip}
\end{equation}
In this case, there is neither a $a_{\text{direct}}$ nor a $a_{\text{rotation}}$ due to the lack of grip at the bottom of the sphere, causing full rotation and no translation.
The next evaluation is the combination of the two previous systems (with and without friction), introducing a variable friction force.
Figure~\ref{fig:PhyiscalPushingFinal} depicts this concept.
The horizontal force $\V F_s$, which arises from $\V F_p$, is now countered by the frictional force $\V F_f$, resulting in the friction between the pole end and the ground.
This force does not necessarily counter $\V F_s$ completely; it depends on the static friction coefficient $\mu_{s\text{Pole}}$.
Therefore 
\begin{equation}
    F_f = \mu_{s\text{Pole}} \cdot F_s = \mu_{s\text{Pole}} \cdot \sin(\zeta) \cdot F_p  \,.
    \label{Eq:F_FPushing}
\end{equation}
The lever force $F_e$ at the end of the pole is now determined from the combination of both leading to
\begin{equation}
 F_e = \cos(\zeta)\cdot (F_s- F_f) = \cos(\zeta)\cdot F_s(1- \mu_{s\text{Pole}})\,.
 \label{Eq:F_eWithFricitionCoeffcient}
\end{equation}
The force $\V F_f$ also acts at the middle of the sphere, just like in the evaluation with the obstacle.
Similarly, it is countered by the frictional force at the contact point of the sphere and ground $\V F_f'$.
This depends now on $\mu_rs$ as the rotational resistance is applied here, which leads to 
\begin{equation}
    F_f' = \mu_{rs} F_f = \mu_{rs}\cdot \mu_{s\text{Pole}} \cdot F_s \,.
    \label{Eq:F_f2}
\end{equation}
The difference $F_f-F_f'$ results in $a_{\text{direct}}$ like in Equation~\eqref{Eq:aDirectObstacle},
\begin{equation}
    a_{\text{direct}}= \frac{F_f-F_f'}{m} = \frac{ (1-\mu_{rs})\cdot \mu_{s\text{Pole}} \cdot F_s}{m}
\end{equation}
Also, $a_{\text{rotation}}$ applies like in Equation~\eqref{Eq:aRotationObstacle}.
However, in this case, the $\omega$ differs as there is not only the torque $\tau_f$ from Equation~\eqref{Eq:tauF} or $\tau_e$ from Equation~\eqref{Eq:tauF} but in fact both.
Let $\tau_{fe}$ be the overall torque working on the sphere. Using Equations~\ref{Eq:F_eWithFricitionCoeffcient} and~\ref{Eq:F_f2}, we get 
\begin{align}
    \tau_{fe} &= \tau_{f'}+\tau_{e} = F_f'\cdot r_m + F_e \cdot  \frac{\cos(\zeta)}{r_m} \\
              &= \mu_{rs}\cdot \mu_{s\text{Pole}} \cdot F_s \cdot r_m + \cos(\zeta)^2\frac{F_s(1- \mu_{s\text{Pole}})}{r_m}   \,.
\end{align}
This then produces a change in angular velocity,
\begin{equation}
    \dot{\omega}= \frac{\tau_{fe}}{I}\,.
\end{equation}
In this case, $a_{\text{rotation}}$ is not merely taken as the surpassed scope of the robot as the whole length is not directly translated to the ground.
As $\mu_{rs}=0$ means just slipping with no rotation being transferred to translation, and $\mu_{rs}=1$ means that the whole scope is also reached as translation, we take the friction coefficient as a linear transition between these two points.
This is an approximation as the real relation between the friction coefficient, rolling resistance coefficient, and the generated horizontal linear acceleration from rotation may not be completely linear. Nonetheless, for the precision aimed at this point, it is still approximated as linear \cite{clover1998longitudinal}.
Therefore,  $a_{\text{rotation}}$ becomes
\begin{align}
    a_{\text{rotation}} &= \mu_{rs}\cdot 2\pi\cdot r_m \cdot \frac{\tau_{fe}}{I} \nonumber \\
    &= \mu_{rs} \cdot 2 \pi\cdot r_m^2 \cdot F_s \cdot I^{-1} \cdot\Bigg(\mu_{rs}\cdot \mu_{s\text{Pole}}   \nonumber \\
    &\quad+ cos(\zeta)^2\frac{(1- \mu_{s\text{Pole}})}{r_m^2}\Bigg)
    \label{Eq:aByRotationPushing}
\end{align}
The overall system is the combination of the acceleration of translation ($a$) and rotation $\dot{\omega}$, which give
\begin{align}
&\begin{bmatrix}
    a \\ \dot{\omega}
\end{bmatrix}   = 
\begin{bmatrix}
    a_{\text{rotation}}+a_{\text{direct}}\\
   \frac{\tau_{fe}}{I}
\end{bmatrix}
    = \\
&\begin{bmatrix}
   \mu_{rs} \cdot 2 \pi \cdot F_s \cdot I^{-1} \cdot(\mu_{rs}\cdot \mu_{s\text{Pole}}\cdot r_m^2\\  + cos(\zeta)^2(1- \mu_{s\text{Pole}})) +\frac{ (1-\mu_{rs})\cdot \mu_{s\text{Pole}} \cdot F_s}{m}\\
   F_s \cdot I^{-1} \left(\mu_{rs}\cdot \mu_{s\text{Pole}} \cdot r_m + \frac{\cos(\zeta)^2(1- \mu_{s\text{Pole}}) }{r_m }\right)
\end{bmatrix}\,.
\label{Eq:aDotOmegaComplete}
\end{align}
Inserting $F_s$ from Equation~\eqref{Eq:F_pSplitUp} leads to
\begin{align}
    \begin{aligned}
    &\begin{bmatrix}
        a \\ \dot{\omega}
    \end{bmatrix}= \sin(\zeta) F_p I^{-1} \\
    & \begin{bmatrix}
 \mu_{rs} \cdot 2 \pi \cdot\Big(\mu_{rs}\cdot \mu_{s\text{Pole}}\cdot r_m^2  \\
    \qquad+ cos(\zeta)^2(1- \mu_{s\text{Pole}})\Big)+\frac{(1-\mu_{rs})\cdot \mu_{s\text{Pole}}}{m}\\
   \mu_{rs}\cdot \mu_{s\text{Pole}} \cdot r_m + \frac{\cos(\zeta)^2\cdot(1- \mu_{s\text{Pole}})}{r_m }
    \end{bmatrix}\,.    
    \end{aligned}
    \label{Eq:AandOmegaWFinalWithoutSplitVH}
\end{align}
For now, we only looked at the horizontal component of $a$ as the gravitational force counters the vertical force.
There certainly exists the possibility that the vertical force exceeds the gravitational force.
In that case, a vertical, linear acceleration will take place, driven by the resulting force.
The resulting vertical force is not considered negative as the force difference between the vertical initiated and gravitational force is countered by the structural force $F_s$.
Let $a_{\text{v}}$ be the vertical acceleration of the robot and $\Theta(x)$ the unit step size function, then
\begin{align}
     a_{\text{v}} &=
     \frac{\Theta(F_R-F_G)(F_R-F_G)}{m} \nonumber \\&= \frac{\Theta (\cos ( \zeta )\cdot F_p-m \cdot g)(\cos ( \zeta )\cdot F_p- m \cdot g)}{m} \nonumber \\
    &= \Theta(\cos ( \zeta )\cdot F_p /m - g)(\cos ( \zeta )\cdot F_p /m-  g)\,.
     \label{Eq:Avertial}
\end{align}
Referring to $a$ in Equation~\eqref{Eq:AandOmegaWFinalWithoutSplitVH} as $a_{\text{h}}$ as it is the horizontal acceleration and inserting Equation~\eqref{Eq:Avertial} give 
\begin{align}
    \resizebox{0.425\textwidth}{!}{$
    \begin{aligned}
        &\begin{bmatrix}
        a_{\text{v}}\\a_{\text{h}} \\ \dot{\omega}
        \end{bmatrix}
        = 
        \begin{bmatrix}
       \Theta(\cos ( \zeta )\cdot F_p /m - g)(\cos ( \zeta )\cdot F_p /m-  g) \\ 0 \\ 0
       \end{bmatrix} \\
       &+\sin(\zeta)F_p  I^{-1}\begin{bmatrix} 0\\
     2 \pi \mu_{rs}(\mu_{rs}\cdot \mu_{s\text{Pole}}\cdot r_m^2  + cos(\zeta)^2(1- \mu_{s\text{Pole}}) )\\
         \qquad\qquad\quad+\frac{ (1-\mu_{rs})\cdot \mu_{s\text{Pole}}  }{m}\\
       \mu_{rs} \mu_{s\text{Pole}} \cdot r_m +  \frac{\cos(\zeta)^2\cdot(1- \mu_{s\text{Pole}})}{r_m}
        \end{bmatrix}\,,
    \end{aligned}
    $}
    \label{Eq:AandOmegaWFinal}
\end{align}

which is the final solution for the introduced model using the pushing approach for locomotion.
We can derive the following trivial behaviors:
\begin{itemize}
    \item Setting $\mu_{rs}$ and $\mu_{s\text{Pole}}$ to one, i.e., full friction and no slip, gives the Equation~\eqref{Eq:aDotOmegaNoSlip}.
     \item Setting $\mu_{rs}$ and $\mu_{s\text{Pole}}$ to zero, i.e., no friction and only slip, gives the Equation~\eqref{Eq:aDotOmegaWithSlip}.
     \item At a $\zeta$ of \unit[0]{rad}, there is neither rotation nor horizontal translation but only vertical movement as we push straight upwards.
     \item $\mu_{rs}$ and $\mu_{s\text{Pole}}$ play a crucial role as they influence most parts of the final system representation as square.
\end{itemize}

\subsection{Leverage}
\label{subsec:Physicalleverage}
\begin{figure}
    \centering
    \includegraphics[width = 0.7 \linewidth]{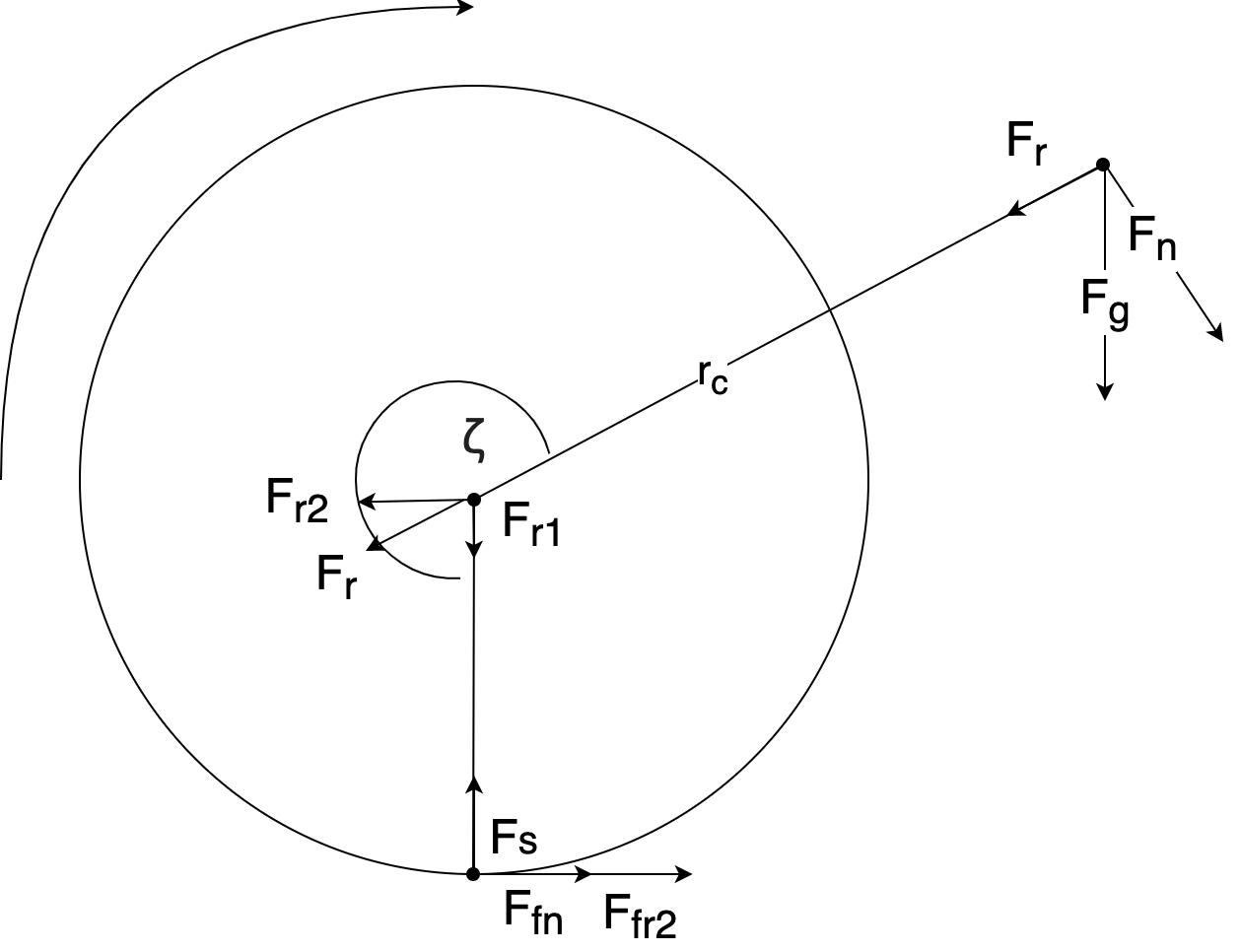}
    \caption{Force evaluation of the leverage approach.}
    \label{fig:PhyiscalLeverage}
\end{figure}

The last part is the physical evaluation of the leverage approach.
Figure~\ref{fig:PhyiscalLeverage} shows the acting forces.
All extended poles are reduced to a point mass at which the gravitational force $\V F_g$ acts.
This force is split up into a normal force $F_n$, perpendicular to the axis of midpoint and center of mass point (with only one pole, this axis is identical with the pole axis), and the force $F_r$, perpendicular to it.
The splitting is done as follows:
\begin{equation}
   \V F_g =  F_g \cdot
    \begin{bmatrix}
    -\cos(\zeta) \\ -\sin(\zeta)
    \end{bmatrix} =  
    \begin{bmatrix}
     F_r \\  F_n
    \end{bmatrix}\,.
\end{equation}
Let $\V \tau_n$ be the torque, initiated by $\V F_n$, and $\V r_c$ be the vector from the center of the sphere to the single mass point of the poles, then
\begin{align}
\V \tau_n = \V r_c \times     \V F_n \nonumber \\
\tau_n = r_c \cdot F_n \,.
\end{align}
However, this is not the only acting torque.
$\V F_r$ acts at the midpoint of the sphere and is split into $\V F_{r1}$ and $\V F_{r2}$ by
\begin{equation}
   \V F_r =  F_r \cdot
    \begin{bmatrix}
    -\cos(\zeta) \\ -\sin(\zeta)
    \end{bmatrix} =  
    \begin{bmatrix}
     F_{r1} \\  F_{r2}
    \end{bmatrix}
    \,.
\end{equation}
$F_{r2}$ introduces a frictional force $F_{f_r2}$ at the bottom of the sphere, which adds another torque to the sphere.
Let $\tau_{fr2}$ be the force introduced by the frictional force $F_{f_r2}$, then
\begin{align}
\tau_{fr2} &= r_m \cdot F_{fr2} =  r_m \cdot \mu_{rs} \cdot F_{r2} \nonumber \\ &= r_m \cdot \mu_{rs} \cdot \sin(\zeta)\cos(\zeta) F_g = r_m \cdot \mu_{rs} \cdot \frac{\sin(2\zeta)}{2} F_g   \,.
\label{Eq:TauFR2}
\end{align}
The difference between the two torques is the resulting torque $\tau_r$,
\begin{align}
    \tau_r &=\tau_n - \tau_{fr2} \nonumber \\ &= r_c \cdot (-\sin(\zeta)) \cdot F_g - r_m \cdot \mu_{rs} \cdot \sin(\zeta)\cos(\zeta)  F_g \nonumber \\ &=  - F_g\left(r_c \cdot \sin(\zeta)  + r_m \cdot \mu_{rs} \cdot \sin(\zeta)\cos(\zeta) \right)\,.
    \end{align}
We can prove that the leverage algorithm produces no torque in the opposite direction of the intended movement because of the following:
\begin{itemize}
    \item $r_c > r_m > 0$
    \item $\pi > \zeta < 2\pi $ leads to $\sin(\zeta) < \sin(\zeta)\cos(\zeta)$ and $\sin(\zeta)<0$
    \item $0 \geq \mu_{rs} \leq 1$.
\end{itemize}
Therefore, $r_c \cdot \sin(\zeta)$ will be negative and always be smaller than $r_m \cdot \mu_{rs} \cdot \sin(\zeta)\cos(\zeta) $, which leads with the multiplication of  $-F_g$ to an overall positive torque.
The initiated rotational acceleration is therefore given by 
\begin{equation}
    \dot{\omega}= \frac{\tau_r}{I} =  - \sin(\zeta) \cdot F_g \cdot I ^{-1}\cdot \left(r_c   + r_m \cdot \mu_{rs} \cdot \cos(\zeta) \right)\,.
\end{equation}
$F_{r1}$ is directly countered by the structural force of the robot.
The part of $F_{r2}$, which is not countered by the friction force, causes direct horizontal translation.
This acceleration is negative, as the acceleration is pointed in the opposite direction then intended.
\begin{align}
    a_{h\text{direct}} &= -\frac{F_{r2}-\mu_{rs}F_{r2}}{m_{\textnormal{robot}}}= 
    -\frac{\sin(\zeta)\cos(\zeta) F_g (1-\mu_{rs})}{m_{\textnormal{robot}}} \nonumber\\ &= \sin(\zeta)\cos(\zeta) g (\mu_{rs}-1) \cdot \frac{m_{\textnormal{lever}}}{m_{\textnormal{robot}}}
\end{align}
Also, the rotation initiates a horizontal translation, depending on $\mu_{rs}$ like in Equation~\eqref{Eq:aByRotationPushing}, with
\begin{align}
    &a_{h\text{rotation}} =\mu_{rs}\cdot 2 \pi \cdot{\omega} \nonumber \\ & = \mu_{rs}\cdot 2 \pi \cdot (-\sin(\zeta) )\frac{F_g}{I}\cdot \left(r_c + r_m \cdot \mu_{rs} \cdot \cos(\zeta) \right)
\end{align}
With the leverage approach for locomotion, there is no possibility for a vertical acceleration, setting $a_v$ to zero.
The moment of inertia $I$ is again taken as constant, as with the pushing approach, for simplification.
Depending on the configuration of the robot and the weight and length of the poles, the $I$ changes during the movement process in non-negligible ways.
In these cases, $I$ needs to be made dependent on $\zeta$, thus making it the function $I(\zeta)$.
We will stay on $I$ for further evaluation.
With this, we set up the overall system:
\begin{align}
    \resizebox{0.425\textwidth}{!}{$
    \begin{aligned}
        \begin{bmatrix}
            a_{\text{v}}\\a_{\text{h}} \\ \dot{\omega}
        \end{bmatrix}
        &= \sin(\zeta)
        \begin{bmatrix}
           0 \\
           \Big(\cos(\zeta) \cdot g \cdot (\mu_{rs}-1)\cdot \frac{m_{\textnormal{lever}}}{m_{\textnormal{robot}}}\\
            - \frac{\mu_{rs}\cdot 2 \pi \cdot F_g}{I}\left(r_c + r_m \cdot \mu_{rs} \cdot \cos(\zeta) \right)\Big)
           \\
           - \frac{F_g}{I}\cdot \left(r_c   + r_m \cdot \mu_{rs} \cdot \cos(\zeta) \right)
        \end{bmatrix}\\
        &=\sin(\zeta)\begin{bmatrix}
           0 \\
           g \cdot m_{\textnormal{lever}}  \cdot \Big(\frac{\cos(\zeta) (\mu_{rs}-1)}{m_{\text{robot}}} \\
           - \frac{\mu_{rs} 2 \pi}{I}\cdot \left(r_c + r_m \cdot \mu_{rs} \cdot \cos(\zeta) \Big)\right)
           \\
           - \frac{F_g}{I}\cdot \left(r_c   + r_m \cdot \mu_{rs} \cdot \cos(\zeta) \right)
        \end{bmatrix}
        \,.
    \end{aligned} 
    $}
    \label{Eq:AandOmegaWFinalLeverage}
\end{align}
It is evident that $\dot{\omega}$ is always positive (ignoring the trivial solutions for $r_c=r_m= 0$, etc.) if $\zeta$ is between \unit[$\pi$]{rad} and \unit[$2\pi$]{rad}.
For the horizontal velocity, determining this is not trivial as for a $\zeta$ between \unit[$\pi$]{rad} and \unit[$2\pi$]{rad}, $a_{h\text{rotational}}$ and $a_{h\text{direct}}$ are in opposite directions.
Therefore, we evaluate
\begin{align}
    0 &\leq a_{\textnormal{h}} \nonumber \\
     0 &\leq \sin(\zeta)\cdot g \cdot m_{\textnormal{lever}} \cdot \left(\cos(\zeta) (\mu_{rs}-1)\cdot \frac{1}{m_{\textnormal{robot}}} 
    \right. \nonumber \\  &  -   \mu_{rs}\cdot 2 \pi\cdot I ^{-1}\cdot \left(r_c + r_m \cdot \mu_{rs} \cdot \cos(\zeta) \right)\bigg)\,.
     \label{Eq:evaluatePositiveALeverage}
\end{align}
We reduce $\sin(\zeta)$ as it is always negative for the evaluated range of $\zeta$, and $g\cdot m_{\textnormal{lever}}$ is also always positive, leading to
\begin{align}
     0 \geq \cos(\zeta) (\mu_{rs}-1) &\cdot \frac{1}{m_{\textnormal{robot}}} - \mu_{rs}\cdot 2 \pi \cdot I ^{-1} \nonumber \\& \qquad\cdot  \left(r_c + r_m \cdot \mu_{rs} \cdot \cos(\zeta) \right)\nonumber \\
    \Leftrightarrow \mu_{rs}\cdot 2 \pi \cdot I ^{-1} &\cdot \left(r_c + r_m \cdot \mu_{rs} \cdot \cos(\zeta) \right) \nonumber \\ &\geq \cos(\zeta) (\mu_{rs}-1) \cdot \frac{1}{m_{\textnormal{robot}}}  \,.
    \label{Eq:testPotiveA}
\end{align}
If we set $\mu_{rs} = 1$, Equation~\eqref{Eq:testPotiveA} becomes 
\begin{align}
     2 \pi\cdot I ^{-1}\cdot \left(r_c + r_m \cdot \cos(\zeta) \right) \geq 0  \nonumber \,.
     \end{align}
Considering that $  2 \pi \cdot I ^{-1}>0$ and that $r_c$ is $r_m$ plus a length $r_x$ dependent on the pole length and the mass distribution of the poles, we further simplify:
     \begin{align}
     & \left(r_c + r_m \cdot \cos(\zeta) \right) \geq 0  \nonumber \\
     \Leftrightarrow&\cos(\zeta) \geq \frac{-r_c}{r_m}  \nonumber \\
     \Rightarrow&\cos(\zeta) \geq   -1 > \frac{-(r_m+r_x)}{r_m} \nonumber \\
     \Rightarrow&\cos(\zeta) \geq  -1
     \,.
\end{align}
This statement will always be true.
Therefore, Equation~\eqref{Eq:evaluatePositiveALeverage} is always fulfilled for $\mu_{rs} =1$.
For the proof that the direction is not always in the direction of desired movement, we use Equation~\eqref{Eq:testPotiveA} and set $\mu_{rs}=0$, which gives
\begin{align}
    0\cdot \frac{2\pi}{I}\cdot \left(r_c + r_m \cdot 0 \cdot \cos(\zeta) \right) &\geq \cos(\zeta) \frac{(0-1)}{m_{\text{robot}}} \nonumber \\ 
    \Rightarrow 0 &\geq -\frac{\cos(\zeta)}{m_{\text{robot}}} \,.
      \label{eq:y0PositiveTranslatioLeverage}
\end{align}
This is true for $\zeta\geq \unit[1.5\pi]{rad}$, as $\frac{1}{m_{\textnormal{robot}}} > 0$.
For $\zeta>\unit[1.5\pi]{rad}$, $F_{r2}$ changes its sign, leading to acceleration in the intended direction, therefore an overall acceleration in the opposite direction is not possible.
Still, this leaves the range of $\unit[\pi]{rad}<\zeta<\unit[1.5\pi]{rad}$.
Therefore, we conclude that not for every value of $\mu_{rs}$, $a_{\text{h}}$ goes in the intended direction of rotation.

Rearranging Equation~\eqref{Eq:testPotiveA} for $\mu_{rs}$, and solving  numerically the worst case of $\zeta$ gives $\zeta=\unit[\pi]{rad}$.
This means at $\zeta=\unit[\pi]{rad}$ we need to evaluate which $\mu_{rs}$ is required to always have translation int the right side by leverage.
After the rearrangement of Equation~\eqref{Eq:testPotiveA} and inserting $\zeta=\unit[\pi]{rad}$, we get 
\begin{align}
    \resizebox{0.425\textwidth}{!}{$
    \begin{aligned}
        \mu_{rs} \geq 
        & \left(\frac{2\pi}{I} \cdot m_{\text{robot}} \cdot \left(\left[\left(\left(\frac{2\pi}{I}\cdot m_{\text{robot}} \cdot  r_c\right)^2\right.\right.\right.\right.\\
        &\qquad\qquad\qquad\qquad+2\cdot \frac{2\pi}{I} \cdot m_{\text{robot}}\cdot r_c \\
        &\qquad\qquad\qquad\qquad\left.+4\cdot \frac{2\pi}{I} \cdot m_{\text{robot}} \cdot r_m +1\right)\\
        &\qquad\qquad\qquad\qquad\left.\left(\frac{2\pi}{I}\cdot m_{\text{robot}}\right)^{-2}\right]^{1/2}+ r_c \Bigg) +1\Bigg)\\
        &\left(2\cdot \frac{2\pi}{I} \cdot m_{\text{robot}}\cdot r_m\right)^{-1}\,.
    \end{aligned}
    $}
\end{align}
Using the moment of inertia of a solid ball around the rolling axis, assuming a straight path, $I= \frac{2}{5}\cdot m_{\text{robot}} \cdot r_m^2$ and some values in the same order of magnitude as the later-introduced prototype $r_m=0.4$, $r_c=0.9$, $m_{\text{robot}}=25$, and $m=0.1$, leads to a minimum $\mu_{rs}$ of $0.012$.
This means that for a guaranteed translation in the desired direction at all times, we need a friction coefficient and rolling resistance coefficient of $0.012$ in total, which is in the magnitude of a ground of ice.

So we conclude that if the sphere experiences no slip, the leverage approach will always generate translation in the intended direction.
For each individual prototype, there exists an $\mu_{rs}<1$ for the same statement, but we are only able to make a worst-case assumption for it.

\section*{Conclusions and Future Work}

Rod-driven locomotion is a promising approach for spherical robots in rough terrains.
The paper has investigated the basic dynamics of locomotion using extendable rods.
We investigated the locomotion by pushing and using the leverage of the rods.
For pushing the maximum extendable length of the rods determines if the maximum rotation speed is limited by static or dynamic considerations.
For the locomotion by leverage we showed,  that for every setup there exists a friction coefficient for which the locomotion by leverage is possible.

Overall, all calculations indicate the basic feasibility of such a system.
Once a sufficient prototype is build,  all introduced ground-laying dynamics will be confirmed by experiments.
For the built these calculations will build the basis for dimensioning of components.
Also,  the calculations will be the foundation for locomotion algorithms using the telescopic linear driven rotation.
Needless to say,  a lot of work remains to be done.
Especially for the frictions coefficients a more sufficient experimental determination has to be investigated.
We showed the huge importance and impact of these friction coefficients on the overall behavior.
Also the interplay between the stabilization of the roll to the sides and the forwards motion need more investigation, resulting in a algorithm combining both tasks into one robust algorithm.





\bibliography{scibib}

\bibliographystyle{unsrt}

\section*{Supplementary materials}

A video demonstrating and showing further experiments of the presented approach  is given under the
following URL: \\
\texttt{https://youtu.be/NyrgArI2zKg}

A video of the DAEDALUS study is available under the following URL:\\
\texttt{https://youtu.be/69CrH9vsTTU}


\clearpage

\end{document}